\documentclass{article}
\usepackage{graphicx}
\usepackage{booktabs}
\usepackage{soul}
\usepackage{hyperref}

\title{Symmetry Reduction Enables Model Checking of More Complex Emergent Behaviours of Swarm Navigation Algorithms}

\author{Laura Antu\~na\footnote{Computer Science Department, Universidade Federal de Minas Gerais, Brazil \newline \href{mailto:laura.antuna@dcc.ufmg.br}{laura.antuna@dcc.ufmg.br}}, Dejanira Araiza-Illan\footnote{Computer Science Department, University of Bristol, UK \newline \href{mailto:dejanira.araizaillan@bristol.ac.uk}{dejanira.araizaillan@bristol.ac.uk}}, S\'ergio Campos\footnote{Computer Science Department, Universidade Federal de Minas Gerais, Brazil \newline \href{mailto:scampos@dcc.ufmg.br}{scampos@dcc.ufmg.br}}, Kerstin Eder\footnote{Computer Science Department, University of Bristol, UK \newline \href{mailto:kerstin.eder@bristol.ac.uk}{kerstin.eder@bristol.ac.uk}}}

\begin{document}

\maketitle

\begin{abstract}
The emergent global behaviours of robotic swarms are important for them to achieve their navigation task goals. These emergent behaviours can be verified to assess their correctness, through techniques like model checking. Model checking exhaustively explores all possible behaviours, based on a discrete model of the system, such as a swarm in a grid. A common problem in model checking is the state-space explosion that arises when the states of the model are numerous. We propose a novel implementation of symmetry reduction, in the form of encoding navigation algorithms relatively with respect to a reference, exploiting the symmetrical properties of swarms in grids. We applied the relative encoding to a swarm navigation algorithm, {\em Alpha}, modelled for the NuSMV model checker. A comparison of the state-space and verification results with an absolute (or global) and a relative encoding of the {\em Alpha} algorithm highlights the advantages of our approach, allowing model checking both larger grid sizes and higher numbers of robots, and consequently verifying more complex emergent behaviours. For example, a property was verified for a grid with 3 robots and a maximum allowed size of $8 \times 8$ cells in a global encoding, whereas this size was increased to $16 \times 16$ using a relative encoding. Also, the time to verify a property for a swarm of 3 robots in a $6 \times 6$ grid was reduced from almost 10 hours to only 7 minutes. Our approach is transferable to other swarm navigation algorithms.
\end{abstract}

\section{Introduction}
Robotic swarms consist of a set of robots with simple individual behaviour rules, working together in cooperation to achieve a more complex or emergent final behaviour. Appealing characteristics of swarms are the low cost incurred in producing the robots, which have a simple hardware design, scalability, and fault tolerance~\cite{Dixon:2011}. Examples of their application to real-life tasks include nanorobotics, disaster rescue missions, and mining or agricultural foraging tasks.

The emergent behaviours of a swarm of robots need to be {\em verified}, with respect to safety and liveness requirements~\cite{Rouff:2004,Winfield:2005}, and {\em validated} to determine whether it is fit for purpose in the target environment. Safety requirements are the allowed behaviours of the system, and liveness requirements specify the dynamic behaviours expected to happen during the execution of the system~\cite{Winfield:2005}. Verification methods include testing over the real robotic platforms or in simulation, and formal, such as model checking and theorem proving. In model checking all the possible behaviours of a system are exhaustively explored to determine whether the requirements are satisfied or violated. Model checking has previously been employed to verify robotic swarms~\cite{Dixon:2010,Dixon:2011,Dixon:2012,Juurik:2011,Konur:2012}.

For model checking the system is modelled in a finite-state manner. The continuous space in which the robots in the swarm move represents a challenge, since it can cause an infinite state model, which translates into a state-space explosion problem when this model is used for verification; i.e., the number of states to explore is beyond computational capabilities. The discretization of the continuous space into cells of fixed size ---i.e., a grid--- is a solution that has been applied in swarms, to enable model checking~\cite{Dixon:2012,Kloetzer:2007}. Even with the discretization of the environment into a ``small'' grid (e.g., $4\times 4$ cells), the state-space explosion problem can occur due to the presence of other variables, which results in too many possible configurations of the robots in the grid.

Symmetry reduction techniques have been used to reduce the size of the models in model checking~\cite{Appold:2010,Bosnacki:2000,Clarke:1996,Donaldson:2005,Emerson:2003,Emerson:2005,Gomes:2010,Norris:1996}. 
These techniques compute a subset of representatives of all the states, after the user provides the classification criteria for the grouping. Alternatively, the classification is computed by analysing similarities amongst the states. Our proposed solution to the state-space explosion problem for swarms in a grid is to exploit the symmetry of the configurations of the robots in the grid, implementing symmetry reduction in a novel manner.

In this paper, we explore the vertical and horizontal symmetry in the grid to reduce the size of the finite-state model. We implemented a {\em relative encoding} of a swarm environment model that eliminates symmetrically equivalent states from the state space. The swarm is assumed to be homogeneous; i.e., all the robots are considered identical in capabilities and rank. In the relative encoding, one robot is set as the ``reference'', with a fixed location and direction of motion. The other robots' locations and directions are defined based on this reference. In a global or absolute encoding, if all the robots in the grid are simultaneously rotated in the same direction and shifted horizontally or vertically by the same distance, the robots' new configurations change in location and direction. In a relative encoding, the locations and directions would remain the same since they are encoded relative to the reference robot, resulting in a reduction of the state space of direction and position configurations.

 We applied our approach to the {\em Alpha} swarm algorithm~\cite{Nembrini:2005}, modelled in the NuSMV model checker~\cite{Cimatti:2002}. An abstraction of the {\em Alpha} algorithm is proposed in~\cite{Dixon:2011,Dixon:2012} and verified through model checking against Linear Temporal Logic (LTL) properties. However, further analysis for grids of more than $8 \times 8$ cells and 3 robots~\cite{Dixon:2011,Dixon:2012} could not be performed due to state-space explosion. Verification results for larger grids and number of robots had to be extrapolated. In~\cite{Kouvaros:2015}, a different abstraction for the {\em Alpha} algorithm  is proposed, using parametrized interleaved interpreted systems, a semantics developed for the verification of multi-agent systems. Expressive temporal-epistemic specifications are verified by parametrizing the model on the number of robots. A ``cut-off'' that represents the behaviour of swarms of any size can be identified. Although our solution does not scale as well as~\cite{Kouvaros:2015} in terms of the number of robots, it scales very well with respect to the grid size, and thus complements~\cite{Kouvaros:2015} in that aspect.

Firstly, we modified the models in~\cite{Dixon:2011,Dixon:2012} to be relative. The state reduction allowed us to check the same swarm property for a larger number of robots and grid size.
Secondly, we abstracted the {\em Alpha} algorithm in our own terms, employing an explicit collision avoidance mechanism, and modelled it using the relative encoding approach.
This new abstraction, despite having more variables, was of a reduced order of states compared to the global encoding in~\cite{Dixon:2011,Dixon:2012}.
The reduction allowed us to check the swarm for the same swarm configurations used in~\cite{Dixon:2011,Dixon:2012}, but obtaining different verification results. The encoding of this new abstraction shows the potential of applying the relative concept to other swarm navigation algorithms in the same manner, which we will be exploring in the future.

The structure of this paper is as follows. Section~\ref{s:swarms} introduces grid-based discrete models for swarm robotics. Section~\ref{s:modelchecking} presents an overview of model checking, the state-space explosion problem and related symmetry reduction techniques. Section~\ref{s:relative} presents the relative encoding model. Section~\ref{s:cases} introduces the {\em Alpha} algorithm. In Section~\ref{s:results} we present, compare and discuss the results of the three encodings of the {\em Alpha} algorithm, using the abstraction in~\cite{Dixon:2011,Dixon:2012} and ours. The concluding remarks are presented in Section~\ref{s:conclusions}.

\section{Swarms in Grids}\label{s:swarms}
When modelling navigating algorithms for swarms, a critical aspect is the continuous space in which the robots in the swarm act. A common approach is to discretize this environment into squared cells of the same size, forming a grid. This grid can ``wrap around'', i.e., it works as if it was projected over a sphere.

Another aspect in the modelling of a swarm is the concurrency of its elements. Four main types have been proposed~\cite{Dixon:2012}: \textit{synchrony}, where all robots move at the same time in each step; \textit{strict turn taking}, where only one robot moves at a time, following a strict order; \textit{non-strict turn taking}, where only one robot moves at a time in a random order, but all the robots get the chance to move after a number of steps; and \textit{fair asynchrony}, where robots move at different time in a random order, and the only guarantee is that a robot will always eventually move.

\section{Model Checking and the State-space Explosion Problem} \label{s:modelchecking}

Model checking is a formal verification method. An exhaustive traversal of the reachable states of a discrete model of the system is performed to check the validity of some desired properties, such as liveness and safety. The reachability depends on the allowed transitions from state to state. The properties to be verified are defined using a temporal logic, for example Linear Temporal Logic (LTL). Model checking is fully automatic. Counterexamples can be produced in the case of a property being false, which helps discovering the reason of the failure~\cite{Clarke:2000}.

Model checkers can be either explicit or symbolic, the latter having an internal structure such as a Binary Decision Diagram (BDD) or Boolean functions that implicitly represent the transitions within the states of the system. The BDDs are constructed before the traversal of the model. Explicit-state model checkers traverse the model whilst verifying it, which may lead to running out of memory before finishing the traversal. Symbolic model checkers allow an initial compression of the model, at the cost of an overhead and memory usage before the checking. NuSMV is an open-source symbolic model checker, with its own input language~\cite{Cimatti:2002}.

The number of states and transitions to traverse in the model can cause problems for model checking (called state-space explosion). Different techniques have been incorporated to alleviate this issue, for instance the use of BDDs that led to the branch of symbolic model checking, symmetry reduction, and abstractions of the model to reduce the number of states~\cite{Clarke:2000}.
%at the cost of meaningfulness and detail.

The symmetrical properties of a finite-state model~\cite{Clarke:2000} can be identified and the state space of the model reduced before model checking. For example, a ``static channel diagram'' of a Promela model is computed in~\cite{Donaldson:2005}.
In~\cite{Gomes:2010}, symmetrical components are reduced by hand, by annotating the model with the directive \texttt{TRANS} to eliminate equivalent transitions from state to state. This manual approach is not trivially transferable to reducing the model of a swarm in a grid.

Symmetry reduction can be applied to BDDs~\cite{Clarke:2000}. ``Quotient models'' are proposed in~\cite{Clarke:1996,Emerson:2005}. Automorphisms that preserve the same transition relations in a BDD (or ``orbit relations'') are computed from permutations of the states, and a chosen representative state substitutes all the states in each orbit relation, forming a reduced quotient model instead of the original BDD.

Symmetry reduction techniques have also been applied to explicit-state model checkers. In~\cite{Norris:1996}, a new data type, ``scalarsets'', is added to the input language of a model checker, to create automorphisms and a ``quotient graph'' (representatives of groups of states) whilst traversing the model. Scalarsets are similar to the static channel diagram model in~\cite{Donaldson:2005}. The automorphisms can be computed on-the-fly along with the traversal of the model, as in~\cite{Bosnacki:2000} based on heuristics. The main disadvantage of all these explicit-state and BDD based symmetry reduction methods is that they are applied into the algorithms of the model checker software or BDD computation, which becomes a non-trivial software re-implementation task.

Our approach to avoid the state-space explosion problem in model checking is based on exploiting symmetrical properties of a swarm in a grid. The encoding of the model in a relative manner with respect to a reference point, as opposed to a global or absolute encoding, can be interpreted as the combination of symmetry reduction and abstraction. The proposed approach is analogous to finding orbit relations or representatives in the global model. The relative encoding employs representatives of the global encoding as possible states, i.e., configurations of the robots in the grid.

\section{Relative Encoding of a Swarm in a Grid}\label{s:relative}

In swarms, the focal point is the interaction of the elements through time. With only the swarm's overall behaviour in mind, the swarm moving north or south, with the same distance and orientation between all the robots and environment elements, such as bounds and obstacles, represents essentially the same situation. Furthermore, these two behaviours correspond to the same swarm configuration if the swarm is encoded in a relative manner. A simple relative encoding is to set a robot as the reference, with fixed location and direction, and to base the location of other robots and environment elements on this reference. A grid populated with robots, rotated and shifted horizontally and vertically, results in different values for the direction and position of each robot in a global or absolute encoding. However, in a relative encoding some rotations and shifting motions correspond to the same grid configurations. 

If a model with $r$ robots in a $m \times m$ size grid (locations), with $d$ possible directions, $p$ other robots' variables of domain sizes $v_i, i=1,...,p$, and $q$ global variables of domain sizes $s_j, j=1,...,q$, is globally encoded, the size of the state space to be explored is $(d \times m^2 \times v_1 \times v_2 \times ... \times v_p)^r \times (s_1 \times s_2 \times ... \times s_q)$. In a relative encoding, the reference robot will have fixed location and direction, and the resulting state space will be of size $(v_1 \times v_2 \times ... \times v_p) \times (d \times m^2 \times v_1 \times v_2 \times ... \times v_p)^{r-1} \times (s_1 \times s_2 \times ... \times s_q)$. This corresponds to a reduction of the state space of $d \times m^2$. In practice this bound changes according to the variables used in the relative encoding of an algorithm.

This decrease in the state space would improve the performance of model checking in terms of time and memory usage, when verifying emergent behaviours of the swarm. Moreover, a counterexample in the relative model is equivalent to a class of counterexamples in the global model.

The update of the direction and location of the robots based on the reference robot's location motion must be performed in two situations: when the reference robot makes a move to an adjacent cell, and when it combines the motion with a change in its direction (rotation). In the first case, the equivalent of the reference robot moving in a direction is the other robots moving in the opposite direction. For example, if the reference moves north, the other robots move south instead. By following this update rule, the distance and direction relation between the swarm of robots remains the same. When the reference robot moves to another cell and also rotates, the other robots make a turn to an opposing direction and their locations need to be updated, as summarized in Table~\ref{tab:direction_update}.

\begin{table}[h!]
  \caption{Location and orientation update after the reference robot changed direction}
  \label{tab:direction_update}
  \centering
  \begin{tabular}{|c|c|c|}
  \hline
    Reference's change  & Direction change& Location change \\ \hline 
    $n \rightarrow e $& $n \rightarrow w$, $s \rightarrow e$, $e \rightarrow n$, $w \rightarrow s$ & $x' = m - y$, $y' = x$ \\ \hline
    $n \rightarrow s$ & $n \rightarrow s$, $s \rightarrow n$, $e \rightarrow w$, $w \rightarrow e$ & $x' = m - x$, $y' = m - y$ \\ \hline
    $n \rightarrow w$ & $n \rightarrow e$, $s \rightarrow w$, $e \rightarrow s$, $w \rightarrow n$ & $x' = y$, $y' = m - x$  \\ \hline 
  \end{tabular}
\end{table}

The relative encoding in NuSMV input language has been designed to have the following structure, to facilitate modularity, employing \texttt{MODULE} constructions: \textit{(a)} we used distances between the reference robot and other robots, instead of specific locations; \textit{(b)} we defined and encoded the update of the distance variables in the \texttt{main} module, along with the concurrency mode logic (e.g., a \texttt{turn} variable); and \textit{(c)} we defined and encoded the motion algorithm (next motion, step size, rotations) of each robot in \texttt{robot} modules, which update the values of the variables in the \texttt{main} module. This procedure can be partially automated through a script to create the \texttt{main} module, from selecting the number of robots and concurrency mode, which we implemented to generate the NuSMV code for the experiments in Section~\ref{s:results}. The \texttt{robot} modules are encoded manually, as they depend on the navigation algorithm.

\section{Alpha Algorithm}\label{s:cases}
The {\em Alpha} algorithm has been used as a case study to demonstrate how to verify emergent behaviours in swarms through model checking tools, as in~\cite{Dixon:2010,Dixon:2011,Dixon:2012,Kouvaros:2015}. In the {\em Alpha} algorithm, the robots in the swarm navigate the environment trying to maintain connectivity, defined as a wireless range. This is achieved by the following rules: \textit{(a)} the default movement of a robot is forward, maintaining its current direction. \textit{(b)} When a robot loses connection with another robot, if the remaining number of connected robots is smaller than a value $\alpha$, the robot makes a 180\,$^{\circ}$ turn. \textit{(c)} Every time a robot regains connectivity with another, it performs a random turn.

A critical requirement for a swarm is: {\em All robots shall eventually be connected.} Expressed formally in LTL, this property was proved to be false in~\cite{Dixon:2011,Dixon:2012}.

\section{Results}\label{s:results}
We compared a global encoding of the {\em Alpha} algorithm based on the abstraction proposed in~\cite{Dixon:2011,Dixon:2012}, against a relative encoding of it, both implemented in the input language of NuSMV. Also, we implemented our own abstraction of the {\em Alpha} algorithm from the ground up, which employs more variables than in~\cite{Dixon:2011,Dixon:2012} for the concurrency modes (strict turn taking, non-strict turn taking, and fair asynchrony), encoded in a relative manner. We measured the state-space reduction of the relative models, compared to the global model.

Table~\ref{tab:results} shows the reduction in the reachable states for the {\em Alpha} algorithm, with grid size $8 \times 8$, 3 robots, and $\alpha = 1$. The state space was computed automatically when compiling the models in NuSMV.
Our new abstraction of the {\em Alpha} algorithm also has fewer states than the global one from~\cite{Dixon:2011,Dixon:2012}. We decided not to consider the fully synchronous concurrency mode in the results, since, in the context of the {\em Alpha} algorithm, it allows for behaviours that are incorrect in the real world: the robots would be allowed to ``swap'' their cell locations.

For the strict turn taking concurrency mode, for example, all three encodings contain a variable \texttt{turn} declared in the \texttt{main} module (a global variable), of domain size 3, and variables \texttt{x}, \texttt{y} and \texttt{direction} declared in the \texttt{robot} module (robot variables), of domain sizes 8, 8 and 4, respectively. Both the global abstraction from~\cite{Dixon:2011,Dixon:2012} and its relative encoding contain a \texttt{motion} variable with domain size 2. The relative encoding also contains a global variable \texttt{random}. Our new abstraction includes the global variables \texttt{random\_turn} and \texttt{random\_move}, with domain sizes of 3 and 2, and robot variables \texttt{last\_num\_con}, with domain size 3. By representing $D_v$ as the domain size of variable $v$ we derive the total number of states for each encoding as being
$(D_x \times D_y \times D_{direction} \times D_{motion})^3 \times D_{turn}$ for the global abstraction in~\cite{Dixon:2011,Dixon:2012},
$(D_x \times D_y \times D_{direction} \times D_{motion})^2 \times (D_{motion}) \times D_{turn} \times D_{random}$ for our relative encoding of the same abstraction,
and $(D_x \times D_y \times D_{direction})^2 \times D_{turn} \times D_{rand\_turn} \times D_{rand\_motion} \times (D_{last\_num\_con})^3$ for our new abstraction. The result of those calculations are mirrored by the values shown in Table~\ref{tab:results}.

\begin{table}[h!]
  \caption{State space size (approximate) for different encodings of the {\em Alpha} algorithm}
  \label{tab:results}
  \centering
  {\footnotesize
  \begin{tabular}{|c|c|r|r|r|} \hline
    Concurrency & Statistic & \multicolumn{2}{|c|}{Abstraction from~\cite{Dixon:2011,Dixon:2012}} & Our new abstraction \\  
                &           		  & 	Global 				& Relative &  Relative  \\ \hline
    Fair        & Total States     & $134.2\times 10^6$ & $1.1\times 10^6$ & $31.9\times 10^6$ \\
    asynchrony  & Reachable States & $68.1\times 10^6$ & $0.5\times 10^6$ & $1.6\times 10^6$ \\ \hline
    Strict      & Total States     & $402.7\times 10^6$ & $3.1\times 10^6$ & $31.9\times 10^6$ \\
    turn taking & Reachable States & $1.1\times 10^6$ & $0.2\times 10^6$ & $0.4\times 10^6$ \\ \hline
    Non-strict  & Total States     & $2818.5\times 10^6$ & $22.0\times 10^6$ & $223.0\times 10^6$ \\
    turn taking & Reachable States & $48.4\times 10^6$ & $0.7\times 10^6$ & $1.3\times 10^6$ \\ \hline
  \end{tabular}}
\end{table}

\begin{figure}[h!]
  \centering
  \includegraphics[width=.75\textwidth]{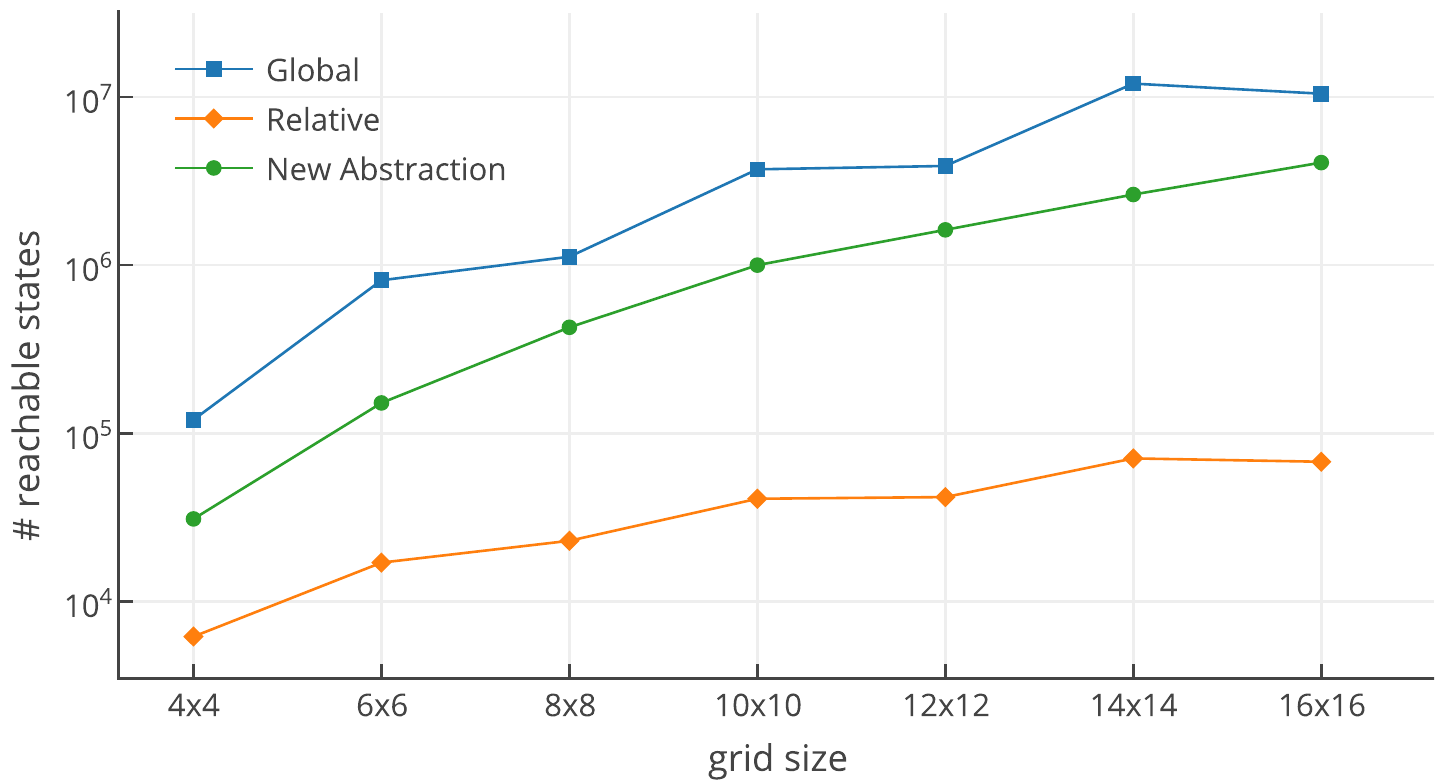}
  \caption{Reachable states as the grid size increases, for strict turn taking concurrency mode and 3 robots}
  \label{fig:grid}
\end{figure}

\begin{figure}[h!]
  \centering
  \includegraphics[width=0.45\textwidth]{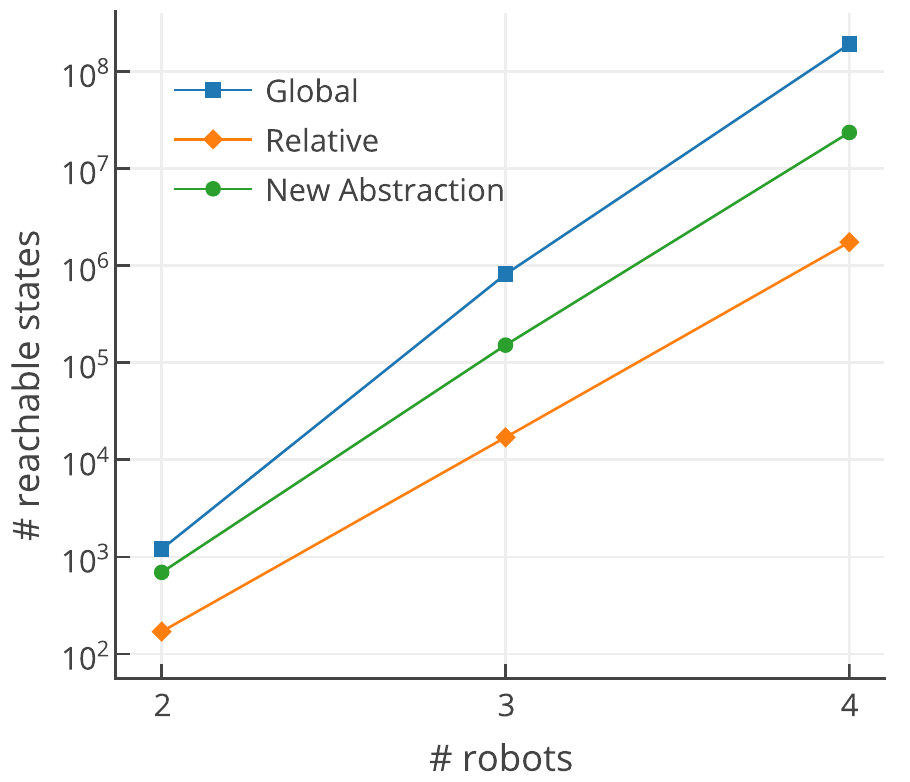}
  \caption{Reachable states as number of robots increases, for strict turn taking concurrency mode and a $6 \times 6$ grid}
  \label{fig:robot}
\end{figure}

Figure~\ref{fig:grid} and Figure~\ref{fig:robot} show a comparison of the state-space for global and relative encodings from the abstraction implemented in~\cite{Dixon:2011,Dixon:2012}, and our new abstraction when varying the grid size or the number of robots, when compiling the models in NuSMV. These experiments consider a strict turn taking concurrency mode. We repeated these experiments for other concurrency modes, non-strict turn taking and fair asynchrony, with similar results in the state-space reduction. In the experiments of Figure~\ref{fig:grid}, the number of robots is set to 3, and the size of the grid is varied. In the experiments of Figure~\ref{fig:robot}, the size of the grid is set to $6 \times 6$, and the number of robots is varied. The final points in the graph correspond to the limit in the memory for the verification to be possible.

We verified the LTL property mentioned previously, {\em all the robots will eventually be connected}, in all these models, to validate the relative encoding of the abstraction in~\cite{Dixon:2011,Dixon:2012}, with respect to their original global model. The verification results were identical, indicating our relative encoding preserved the properties of the global encoding. For the verification, we used NuSMV version 2.5.4, running on a PC with Ubuntu 14.10 and 4\,GB RAM.

\begin{figure}[h!]
  \centering
  \includegraphics[trim=2cm 14.75cm 1cm 2.5cm, clip, width=0.75\textwidth]{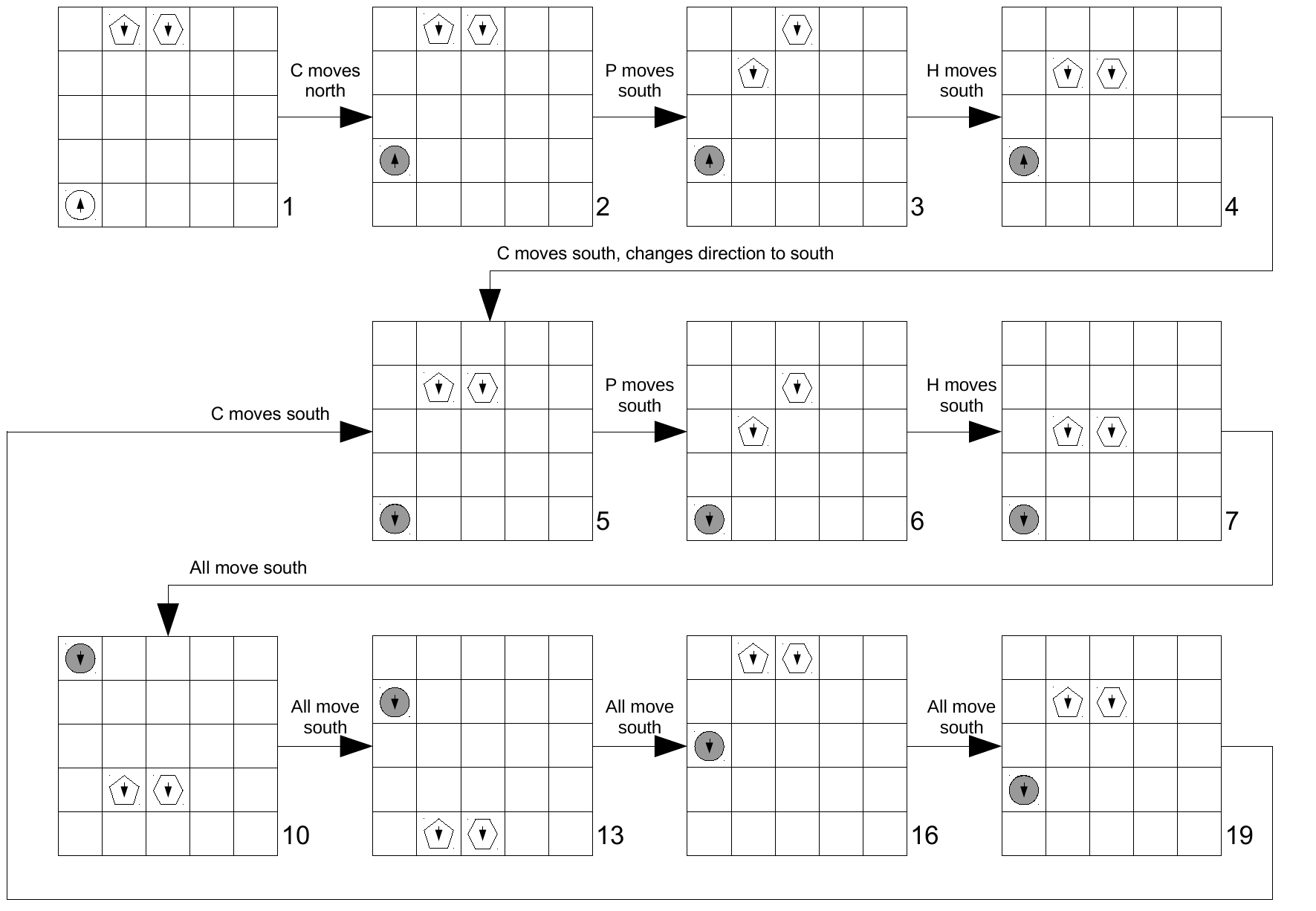}
  \caption{Failing trace of globally encoded model. C: circle, P: pentagon, H: hexagon. Disconnected robots in gray}
  \label{fig:global_trace}
\end{figure}

\begin{figure}[h!]
  \centering
  \includegraphics[trim=2cm 5cm 1cm 16.75cm, clip, width=0.75\textwidth]{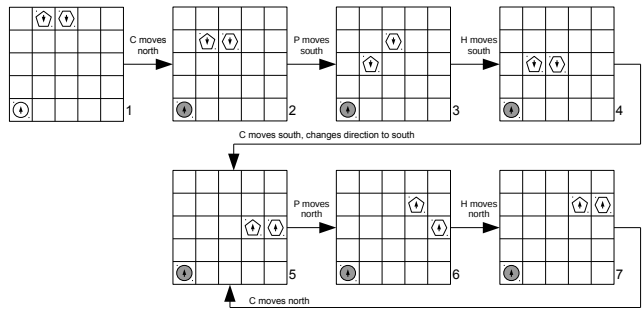}
  \caption{Failing trace of relatively encoded model. C: circle (reference), P: pentagon, H: hexagon. Reference: circle. Disconnected robots in gray}
  \label{fig:relative_trace}
\end{figure}

A counterexample (failing trace) provided by the model checker, with strict turn taking concurrency mode, 3 robots, and a grid of size $5 \times 5$, is shown in Figure~\ref{fig:global_trace}, for the global encoding. From state 5 onwards, all robots move south in a loop and robot C (circle) never reconnects to the swarm. In the global model, it takes 15 states until the loops starts. However, the same pattern is observed within each 3 steps from the moment when robot C (circle) changes its direction. This repetitiveness was eliminated in the relative model, as illustrated by Figure~\ref{fig:relative_trace}, achieving a reduction of 12 states. In the relative encoding, the location of the reference (circle) is fixed to cell $(0,0)$ and its direction to North. Subsequently, other robots update their position and orientation according to the decisions of the reference robot, or according to their individual decisions, as explained in Section~\ref{s:relative}.

Posteriorly, we compared the verification results using different abstractions of the {\em Alpha} algorithm, the relative version of the one proposed in~\cite{Dixon:2011,Dixon:2012} and our new abstraction, also relative. These experiments were conducted for $\alpha=1$, and a strict turn taking concurrency mode. We found that the LTL property is true for some settings, such as 3 robots in grids of $2\times2$, $3\times3$ and $4\times4$, and two robots in a $5\times5$ grid. For other settings, such as 3 robots in a $5\times5$ grid, the property is false and the swarm will not regain connection. This is caused by the randomness of the individual robots decisions, that can be repeated infinitely, i.e., a robot can infinitely often perform the same patterns of motion but they do not lead to regain connectivity.

\begin{figure}[h!]
  \centering
  \includegraphics[width=0.6\textwidth]{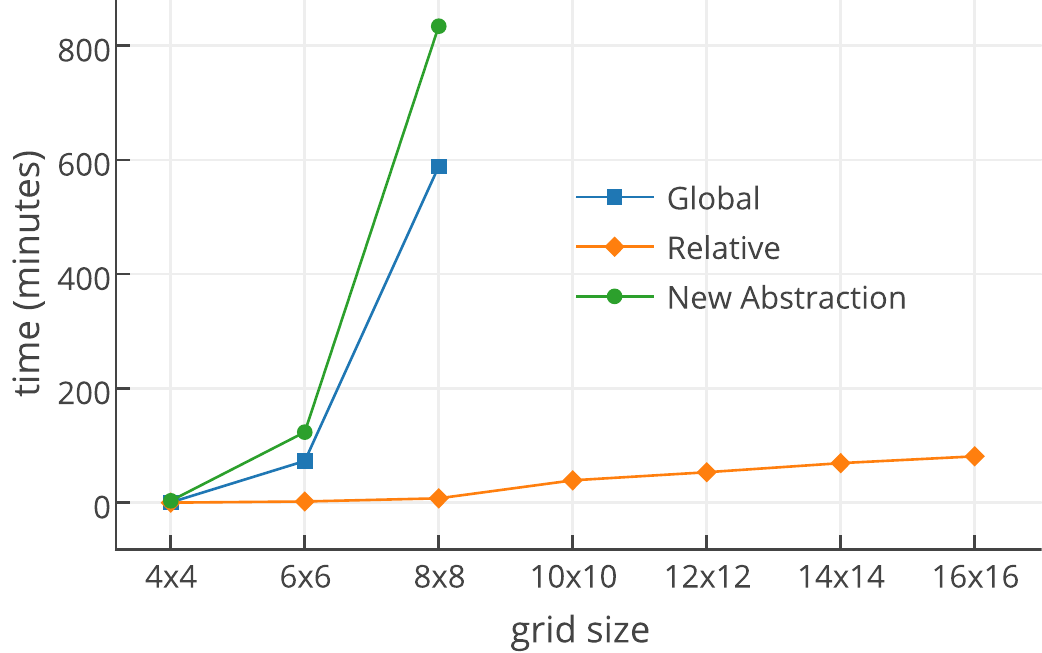}
  \caption{Verification time for the LTL property in Section~\ref{s:cases} as the grid size increases, for strict turn taking concurrency mode and 3 robots}
  \label{fig:timegrid}
\end{figure}

\begin{figure}[h!]
  \centering
  \includegraphics[width=0.35\textwidth]{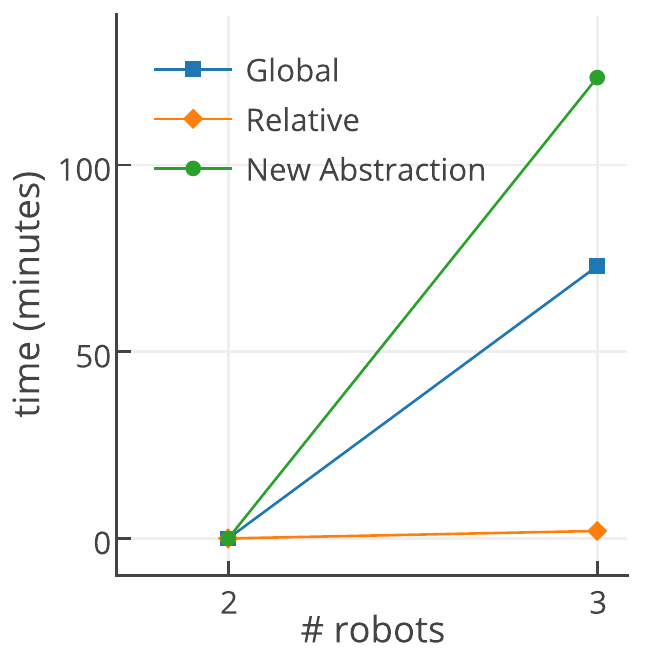}
  \caption{Verification time for the LTL property in Section~\ref{s:cases} as number of robots increases, for strict turn taking concurrency mode and a $6 \times 6$ grid}
  \label{fig:timerobot}
\end{figure}

Figure~\ref{fig:timegrid} and Figure~\ref{fig:timerobot} show the verification time for the LTL property in Section~\ref{s:cases} for global and relative encodings from the abstraction~\cite{Dixon:2011,Dixon:2012}, and the new abstraction when varying the grid size or the number of robots. These experiments, as before, consider a strict turn taking concurrency mode. In the experiments of Figure~\ref{fig:timegrid}, the number of robots is set to 3, and the size of the grid is varied. In the experiments of Figure~\ref{fig:timerobot}, the size of the grid is set to $6 \times 6$, and the number of robots is varied. The final points in the graph correspond to a stipulated time limit of 5 days for the verification.

We observed the same time reduction patterns between the global and relative encodings of the abstraction in~\cite{Dixon:2011,Dixon:2012}, as a consequence of the state-space reduction. The number of robots is a more significant constraint to the verification time than the grid size. Although it was not possible to verify the relative encoding with 4 robots due to time restrictions, applying some constraints to the initial configuration of the swarm allowed the generation of a counterexample, a proof that the property is false for that number of robots and grid size. In contrast, the global encoding could not be verified, even when applying the same constraints. The relative encoding of the new abstraction of the {\em Alpha} algorithm takes longer to be verified than the previous abstraction, as it is more complex. Nevertheless, it allowed us to obtain different verification results compared to~\cite{Dixon:2011,Dixon:2012}, which we believe are closer to the intention of the {\em Alpha} algorithm. The difference between the two abstractions of the {\em Alpha} algorithm need to be further investigated to determine if any are incorrect, given the verification results.

\section{Conclusions}\label{s:conclusions}
We presented an approach that achieves significant state-space reduction and thus allows model checking more complex emergent behaviours of swarms. We propose the use of symmetry reduction to eliminate symmetrical states (i.e., configurations of robots in the grid), implemented through a relative encoding of the swarm, where one robot is the reference, and the others' navigation is encoded relative to it. This encoding, compared to a global one, helps to reduce the state space of the model, as demonstrated by our results in Section~\ref{s:results}. Thus, verification through model checking can be performed over larger grid sizes and higher numbers of robots, and also for more detailed abstractions that model navigation algorithms using more variables. Although the state space reduction is more significant in terms of the grid size, and not as expressive if considering realistic swarm sizes, analysing small robot groups can help to understand larger swarms within the lower limit bounds of swarm size, which demands more from the navigation algorithm~\cite{Dixon:2011,Dixon:2012}.

Future work includes running more experiments with different $\alpha$ values, and the verification of more complex properties of robotic swarms, such as the emergence of teams or different swarms due to the connectivity properties. The successful relative encoding of a new abstraction of the {\em Alpha} algorithm gives us confidence that our approach can be applied to other swarm algorithms. In the future we would like to model navigation algorithms such as the {\em Beta} algorithm~\cite{Nembrini:2005} in a relative manner, to validate the transferability of our approach.

\paragraph{Aknowledgements}
We would like to thank Clare Dixon for providing her {\em Alpha} algorithm models, and her invaluable comments and advice. The work by D.\ Araiza-Illan and K.\ Eder was partially supported by the EPSRC, grants EP/J01205X/1 RIVERAS: Robust Integrated Verification of Autonomous Systems and EP/K006320/1 Trustworthy Robotic Assistants.

\bibliographystyle{plain}
\bibliography{ref}

\end{document}